\documentclass[runningheads]{llncs}

\usepackage{eccv}

\usepackage{eccvabbrv}

\usepackage{graphicx}
\usepackage{booktabs}
\usepackage{algorithm}
\usepackage{algpseudocode}
\usepackage{multirow}
\usepackage{arydshln} % 支持虚线
\usepackage{amsmath} 
\usepackage{diagbox}

\usepackage{colortbl} % 用于给行加底色
\usepackage{xcolor}
\definecolor{grayrow}{gray}{0.95} % 定义淡淡的灰色底色
\usepackage[accsupp]{axessibility}  % Improves PDF readability for those with disabilities.

\usepackage{hyperref}

\usepackage{orcidlink}

\begin{document}

%\title{Enhancing Adversarial Robustness via Structural Invariance and Signed Distance Fields} 
\title{Enhancing Adversarial Robustness with Signed Distance Fields for Harmonizing Geometric Invariance and Texture} 

\titlerunning{Structural Invariance for Robustness}

\author{Zhe Li \and Bernhard Kainz}

\authorrunning{F.~Author et al.}

\institute{Department AIBE, FAU Erlangen-Nürnberg, Erlangen, Germany \\
\email{zhe.li@fau.de}}

\maketitle

\begin{abstract}
Deep neural networks demonstrate impressive performance in visual recognition but remain highly vulnerable to imperceptible adversarial attacks. Existing defense strategies such as adversarial training and diffusion-based purification have achieved significant progress but are frequently constrained by high computational cost, information loss, and inference latency. To address these challenges, we propose a Geometric and Texture balancing Purification (GeoTexPuri) framework that enhances adversarial robustness by harmonizing invariant geometric structures with textural features. Specifically, the framework integrates dense geometric guidance into the training phase by transforming discrete image masks into continuous spatial fields via Signed Distance Fields (SDF). This process establishes stable structural anchors that shield the model from local pixel noise. Through a multi-stream training objective, the model learns to internalize purified representations that effectively align semantic textural cues with these underlying geometric invariants.
Extensive experiments on ImageNet demonstrate the efficacy of our approach. GeoTexPuri achieves 84.79\% clean accuracy and 83.52\% robust accuracy under the AutoAttack. Crucially, GeoTexPuri functions as a deterministic classifier during inference, requiring only the input image without any auxiliary geometric modules or additional computational costs, thereby ensuring a scalable and efficient solution for real-time applications.

\end{abstract}

%-------------------------------------------------------------------------
\section{Introduction}

Modern computer vision models achieve remarkable success across a wide range of complex visual tasks but remain susceptible to adversarial perturbations. The existence of input modifications imperceptible to human observers often triggers catastrophic predictive failures. 
This instability demonstrates that standard neural representations frequently overfit to training distributions and cannot reliably withstand minor distribution shifts introduced by adversarial noise. 
In critical safety domains including autonomous navigation and biometric authentication, the requirement for consistent performance turns these model vulnerabilities into a significant operational risk.
Consequently, the development of robust defense methodologies has become a primary objective for the research community.

\begin{figure}[tb]
  \centering
  \begin{subfigure}{0.235\linewidth}
    \includegraphics[width=\linewidth]{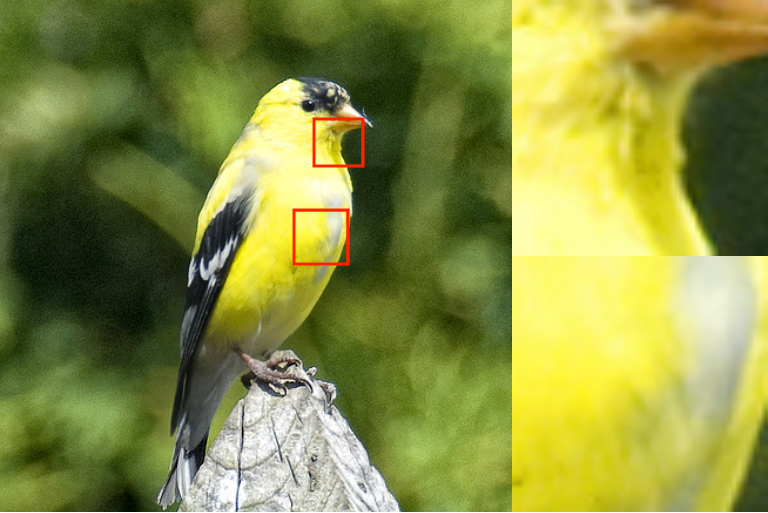}
    \caption{Clean}
  \end{subfigure}
  \begin{subfigure}{0.235\linewidth}
    \includegraphics[width=\linewidth]{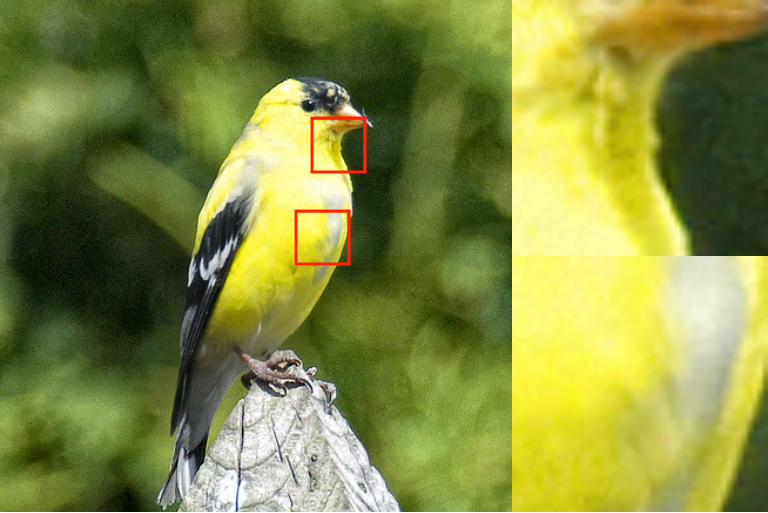}
    \caption{Adversarial Image}
  \end{subfigure}
  \begin{subfigure}{0.235\linewidth}
    \includegraphics[width=\linewidth]{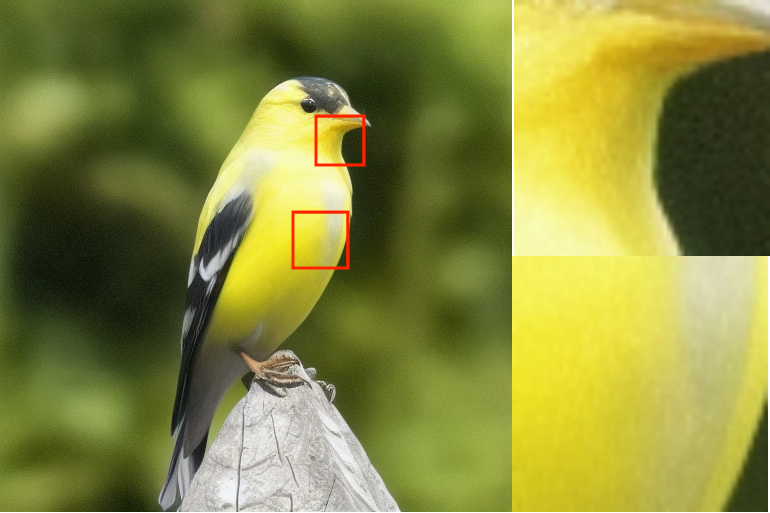}
    \caption{Diffusion-based}
  \end{subfigure}
  \begin{subfigure}{0.235\linewidth}
    \includegraphics[width=\linewidth]{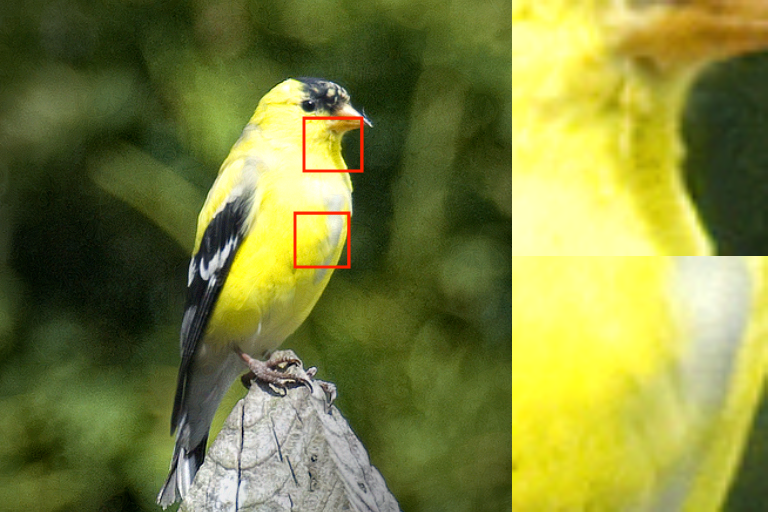}
    \caption{Adv \(\odot\) SDF (Ours)}
  \end{subfigure}
\caption{Illustration the oversmoothing effect of diffusion-based purification. Compared to the clean image, the purified output (c)  loses plumage details and exhibits oversmoothed textures. Our approach (d) preserves these semantic features.}
\label{fig:teaser}
\end{figure}

Existing research in adversarial defense has primarily developed along two paradigms: adversarial training and adversarial purification.
Adversarial training methods involve incorporating adversarially perturbed samples into the training pipeline to bolster model resilience against specific attack budgets. However, these techniques often exhibit a performance trade-off between robustness and clean accuracy and may show limited generalization to unseen or adaptive threats.
Adversarial purification provides an alternative by treating robustness as an inference-time preprocessing task that aims to project adversarial inputs back onto the clean image manifold. Diffusion-based purification \cite{nie2022diffusion,wang2022guided,lei2025instant} has recently emerged as a prominent framework that utilizes iterative denoising to remove perturbations. Despite their demonstrated effectiveness, these generative paradigms suffer from two notable bottlenecks. 
The first is the high computational cost stemming from the multi-step reverse sampling process, which introduces significant latency for real-time applications.
The second concerns the preservation of semantic fidelity.
The generative objective of diffusion models often prioritizes global distribution matching over the preservation of local structural details. This bias leads to the oversmoothing of instance-specific features and the unintended erasure of fine-grained textures. As shown in Fig.~\ref{fig:teaser} (c), the intricate patterns of plumage are replaced with smooth surfaces.
These observations motivate the need for adversarial defenses that suppress perturbations while preserving texture details.

This pursuit requires identifying the root cause of existing vulnerabilities, which is tied to the fact that standard convolutional architectures exhibit an inherent texture bias~\cite{Geirhos19,Ilyas19,hermann2020origins}. Consequently, adversarial attacks can easily manipulate these brittle local pixel-level details to trigger predictive failures. 
Such behavior contrasts with human perception, which relies on global robust shape preferences that remain consistently invariant to pixel-level noise. This intrinsic structural stability suggests that incorporating shape-centric information can effectively bridge the gap between clean classification performance and adversarial robustness.

In this work, we introduce Geometric and Texture balancing Purification
(GeoTexPuri), a deterministic and lightweight defense framework that operationalizes these principles by seeking an equilibrium between appearance and structure. 
Our approach provides explicit geometric guidance to counteract the inherent texture bias of neural networks by augmenting the input with its Signed Distance Field (SDF). This field represents a continuous and low-frequency spatial mapping of object boundaries derived from global topology.
The SDF exhibits remarkable stability under local pixel-level perturbations, which effectively guides the classifier to emphasize structural invariants while retaining indispensable textural details.
Unlike iterative generative models, GeoTexPuri is a diffusion-free purification mechanism that requires no additional processing steps during inference. This ensures optimal defense efficiency and enables its deployment in real-world applications.
Experimental results show that GeoTexPuri achieves superior performance on ImageNet, reaching 83.52\% robust accuracy against AutoAttack and exceeding current state-of-the-art benchmarks by a significant margin of 9.33\%. 
In ablation, we validate the effectiveness of GeoTexPuri across diverse and challenging benchmarks such as object classification on the full-scale ImageNet and face recognition on CelebA-HQ, demonstrating its broad applicability and robustness.

Our main contributions are as follows:

\begin{enumerate}
  \item We propose the Geometric and Texture balancing Purification (GeoTexPuri) framework, a unified training strategy that enhances adversarial robustness by harmonizing invariant geometric structures with textural features. By internalizing purified representations during training, GeoTexPuri avoids the additional computational cost during inference.

  \item We leverage continuous spatial representations via Signed Distance Fields (SDF) to provide dense differentiable geometric guidance. By fusing these structural signals with semantic textural features, the model establishes spatial invariants that effectively shield it from adversarial perturbations.

  \item Extensive evaluations on ImageNet demonstrate that GeoTexPuri significantly outperforms existing defense methods. Our framework achieves 84.79\% clean accuracy and 83.52\% robust accuracy under the AutoAttack protocol.
\end{enumerate}

%----------------------------------------------------------------------
%----------------------------------------------------------------------

\section{Related Work}
\label{sec:related}

\noindent\textbf{Adversarial Training.}
Conventional adversarial training (AT) methods aim to improve model robustness by incorporating adversarially perturbed samples directly into the optimization process~\cite{goodfellow2014explaining}, with Projected Gradient Descent (PGD) \cite{madry2017towards} serving as a primary benchmark.
Recent research has achieved substantial improvements through the utilization of large-scale datasets~\cite{gowal2021improving} and architectural refinements such as the ConvStem in vision transformers~\cite{singh2023revisiting}. 
Furthermore, techniques such as MeanSparse~\cite{amini2024meansparse} leverage feature sparsification performed after training to enhance stability, whereas IJSAT~\cite{lau2023interpolated} introduces interpolated joint spatial and adversarial training to bolster resilience. Although these developments enhance defensive capabilities, they often exhibit a performance trade-off regarding clean accuracy and may manifest limited efficacy beyond specific categories of adversarial attacks.

\noindent\textbf{Adversarial Purification.}
Purification functions as a flexible defense strategy by preprocessing inputs to remove perturbations before they reach the classifier. Early purification strategies utilized generative models, such as Defense GAN~\cite{samangouei2018defense} or ensembles of Variational Autoencoders~\cite{Schott19}, to project perturbed inputs onto the clean image manifold. However, these methods often struggle with scalability and adaptive adversaries. Recently, Diffusion-based denoising has emerged as a dominant paradigm. DiffPure~\cite{nie2022diffusion} utilizes forward and reverse diffusion processes, whereas GDMP~\cite{wang2022guided} and Bai~\etal~\cite{bai2024diffusion} introduce guided reverse sampling to stabilize semantic recovery. Despite their empirical success, the iterative sampling process required by diffusion models introduces substantial computational costs and high inference latency. Although OSCP~\cite{lei2025instant} attempts to reduce this overhead through distillation, the underlying diffusion process remains prone to information loss and the oversmoothing of critical structural details. %This loss of fidelity indicates that generative purification often lacks the necessary constraints to preserve fine-grained features essential for precise classification, especially when evaluated against an ensemble of attacks such as AutoAttack~\cite{croce2020reliable}.

\noindent\textbf{Geometric Priors and Structural Awareness.} 
Previous research utilizes an edge map as a condition for Diffusion-based models~\cite{lei2025instant} to control the purification process. However, this edge representation is inherently sparse and lacks informative gradients across the image domain, which restricts its efficacy in recovering complex semantic content. 
In contrast, continuous geometric fields provide dense structural information that spans the entire spatial domain. 
Current methods for constructing SDF, such as those used in shape representation~\cite{park2019deepsdf,wu2024clusteringsdf} and neural implicit surface reconstruction~\cite{sitzmann2020implicit}, require explicit 3D supervision or multi-view consistency. They are difficult to apply directly to adversarial purification utilizing only 2D images. 
Furthermore, while salient object detection and segmentation techniques extract interest regions through pyramid structures~\cite{kim2022revisiting} or bilateral reference mechanisms~\cite{zheng2024bilateral}, they often lack the ability to distinguish between different object categories within a single scene. To address these limitations, we synergize Grounding DINO~\cite{liu2024grounding} for category identification with HQ-SAM~\cite{ke2023segment} to obtain object masks aligned with labels.
%These masks are then used to construct SDF that provide invariant geometric guidance for the subsequent training process.

%----------------------------------------------------------------------
%----------------------------------------------------------------------
\section{Method}
\begin{figure}[tb]
  \centering
    \includegraphics[width=\linewidth]{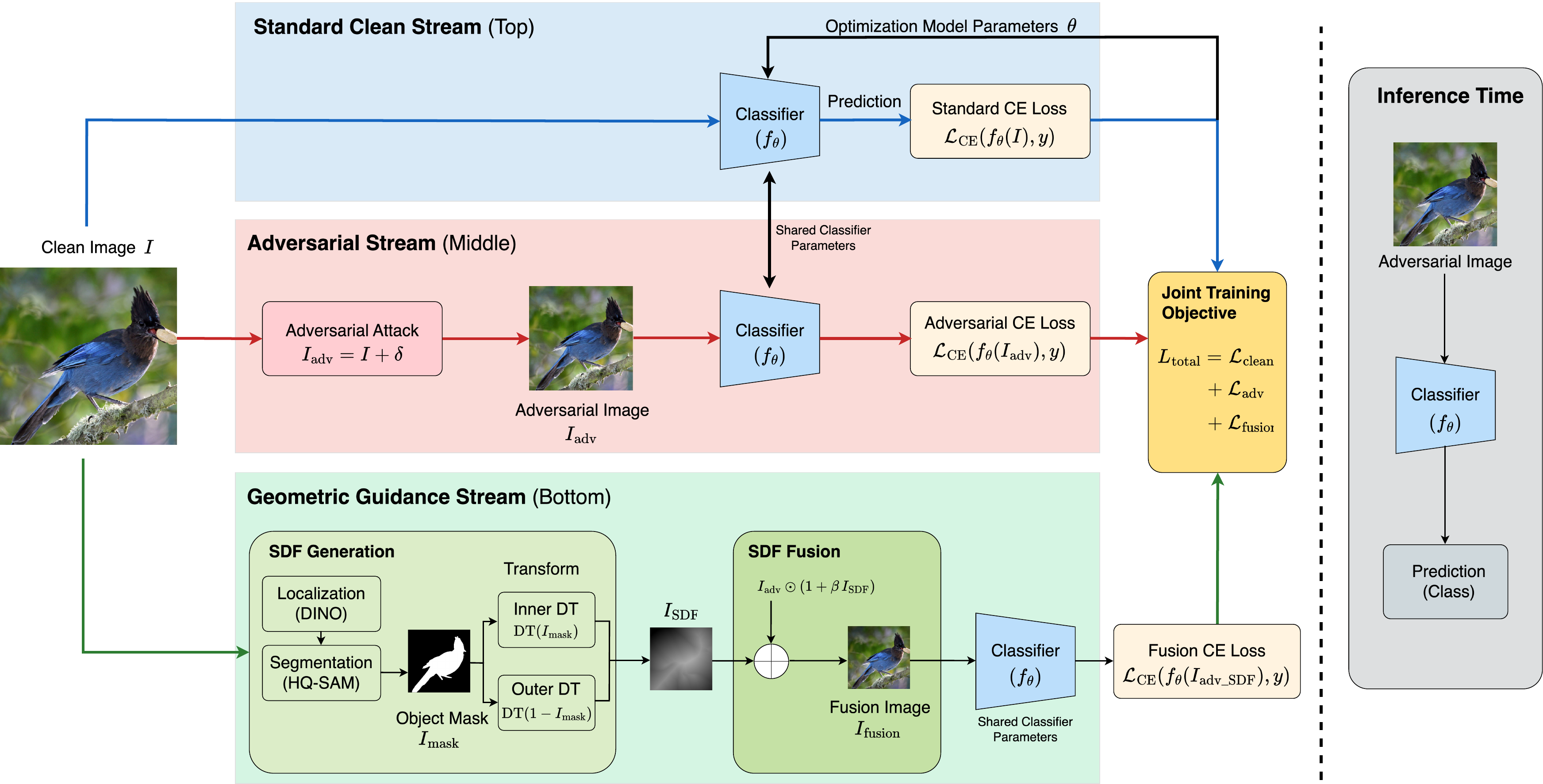}
   \caption{Overview of the GeoTexPuri Framework. 
   The training process (left) employs a multi-stream strategy to harmonize geometric and textural features. (1) The Standard Clean Stream (blue) processes original images to preserve essential semantic features and enhance classification performance on clean data. (2) The Adversarial Stream (red) introduces perturbations via an adversarial attack.
   (3) The Geometric Guidance Stream (green) load stored target object mask and computes SDF. This geometric information is then fused with the adversarial input to provide explicit spatial guidance. The model is optimized using a joint training objective $L_{\mathrm{total}}$ (orange). At Inference Time (gray, right), the model operates as a standard classifier and achieves enhanced adversarial robustness without the SDF.}
   \label{fig:framework}
\end{figure}

GeoTexPuri is a unified training framework designed to enhance adversarial robustness by harmonizing invariant geometric structures with textural features. Our approach counteracts the inherent texture bias of neural networks by integrating geometric information directly into the training phase, which enables the model to internalize purified representations that are robust to appearance perturbations. Given a clean image $I$ and its adversarial counterpart $I_{\mathrm{adv}} = I + \delta$, where $\|\delta\|_p \leq \epsilon$, our objective is to optimize the model parameters $\theta$ such that the classifier $f_{\theta}$ maintains consistent and accurate predictions, i.e., $f_{\theta}(I_{\mathrm{adv}}) \approx f_{\theta}(I)$.
As shown in Fig.~\ref{fig:framework}, GeoTexPuri leverages a multi-streams training strategy to provide diverse learning signals. %Specifically, we incorporate SDF to reinforce geometric awareness and overcomes excessive model reliance on non-robust textures. These streams are optimized through a joint training objective, allowing the network to capture the structural essence of the visual content. 
%We develop an automated pipeline to convert input images into SDF for the target category, which provide invariant geometric cues that are robust to appearance perturbations. 

%----------------------------------------------------------------------
%----------------------------------------------------------------------

\subsection{Preliminaries}

SDF is a scalar function that encodes the distance from any point in space to its closed surface, with the sign indicating whether the point lies inside or outside the surface. Following the classical formulation in level-set methods \cite{osher1988fronts}, we define the SDF as:

\begin{equation} 
\phi(\mathbf{x}) = 
\begin{cases} 
+\min_{\mathbf{y} \in \partial \Omega} \|I - \mathbf{y}\|, & I \in  \Omega \quad (\text{inside}) \\ 
0, & I \in \partial \Omega \quad (\text{on surface}) \\ 
-\min_{\mathbf{y} \in \partial \Omega} \|I - \mathbf{y}\|, & I \in \mathbb{R}^n \setminus \Omega \quad (\text{outside}),  
\end{cases} 
\label{eq:sdf} 
\end{equation}

where $\Omega \subset \mathbb{R}^n$ is a closed domain, $\partial \Omega$ denotes its boundary, and $\|\cdot\|$ is the Euclidean norm. By construction, the zero level set of $\phi(I)$ implicitly defines the surface, the gradient $\nabla \phi(I)$ points outward along the surface normal, and $|\phi(I)|$ gives the exact distance to the surface.

The SDF provides a continuous and differentiable representation of geometry, which is particularly advantageous for gradient-based optimization, collision detection, level set methods, and rendering. Its key properties are as follows. First, the magnitude $|\phi(I)|$ represents the exact Euclidean distance to the nearest surface. Second, the gradient $\nabla \phi(I)$ is a normalized vector pointing outward from the object interior, \emph{i.e.}, $\|\nabla \phi(I)\| = 1$ almost everywhere. The zero level set ($\phi(I) = 0$) implicitly defines the surface geometry, enabling operations such as Boolean combinations, surface smoothing, and volumetric manipulations.

%The geometric properties of the SDF contribute to its notable robustness against local pixel-level perturbations. Unlike raw intensity values that adversarial attacks directly manipulate, the SDF encodes the global topology of object boundaries and is insensitive to local pixel-level changes. Consequently, adversarial noise induces only negligible deviation in the corresponding SDF, even when it is sufficient to corrupt classifier predictions in pixel space. This observation motivates our purification strategy. We fuse the SDF with the adversarial image to provide the classifier with a structurally stable representation. The fused representation preserves the semantically discriminative texture of the original image while anchoring classification decisions to geometric invariants that adversarial perturbations cannot readily distort.

%----------------------------------------------------------------------
%----------------------------------------------------------------------

%\section{Geometric and texture balancing purification framework}

\begin{figure}[tb]
  \centering
  \begin{subfigure}{0.24\linewidth}
    \includegraphics[width=\linewidth]{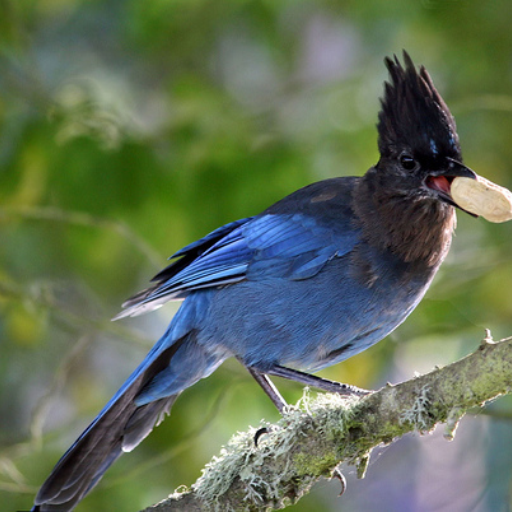}
    \caption{Clean}
  \end{subfigure}
  \begin{subfigure}{0.24\linewidth}
    \includegraphics[width=\linewidth]{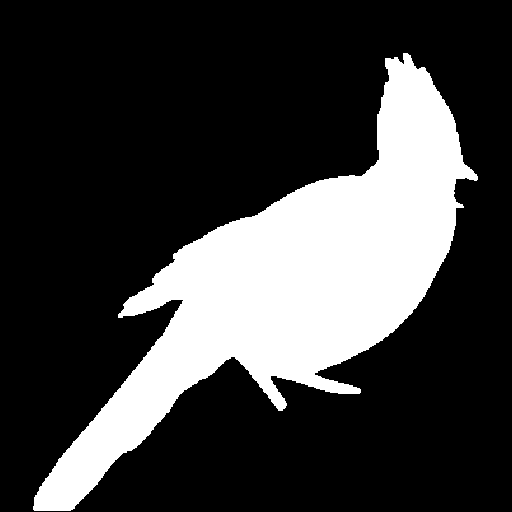}
    \caption{Mask}
  \end{subfigure}
  \begin{subfigure}{0.24\linewidth}
    \includegraphics[width=\linewidth]{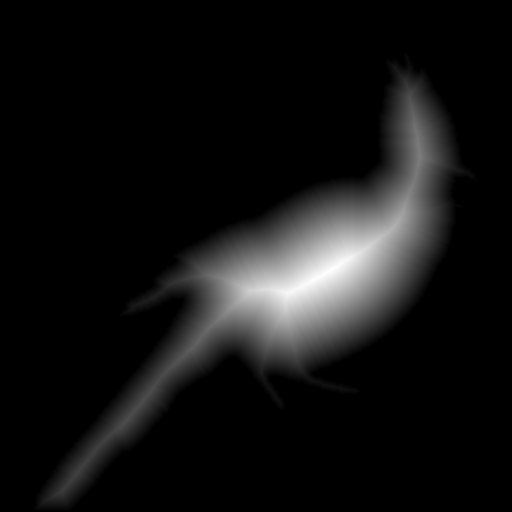}
    \caption{Inner Distance}
  \end{subfigure}
  \begin{subfigure}{0.24\linewidth}
    \includegraphics[width=\linewidth]{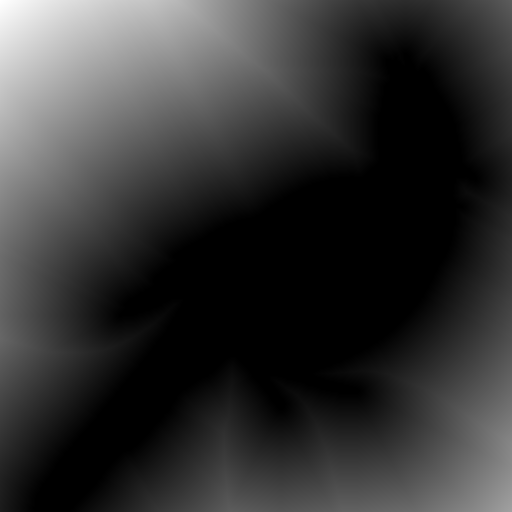}
    \caption{Outer Distance}
  \end{subfigure}
  \begin{subfigure}{0.24\linewidth}
    \includegraphics[width=\linewidth]{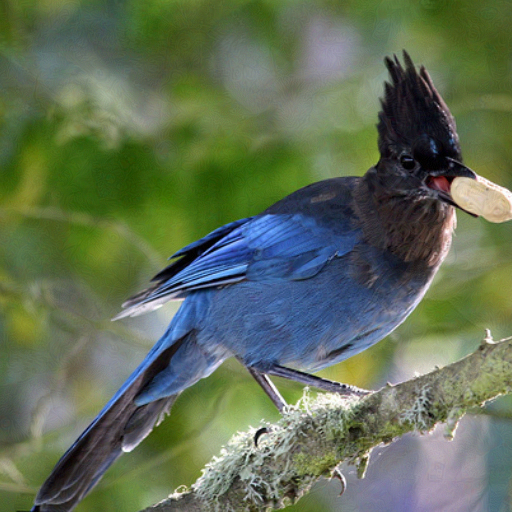}
    \caption{Adversarial Image}
  \end{subfigure}
  \begin{subfigure}{0.24\linewidth}
    \includegraphics[width=\linewidth]{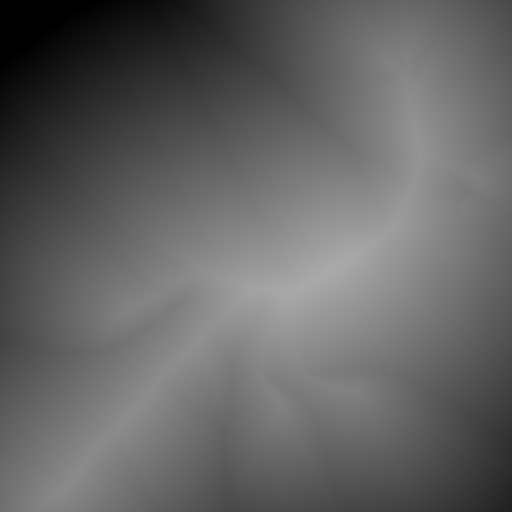}
    \caption{SDF}
  \end{subfigure}
  \begin{subfigure}{0.24\linewidth}
    \includegraphics[width=\linewidth]{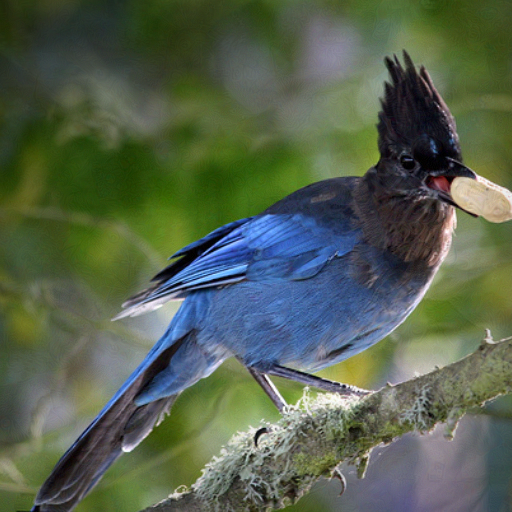}
    \caption{Adv \(\odot\) SDF}
  \end{subfigure}
  \begin{subfigure}{0.24\linewidth}
    \includegraphics[width=\linewidth]{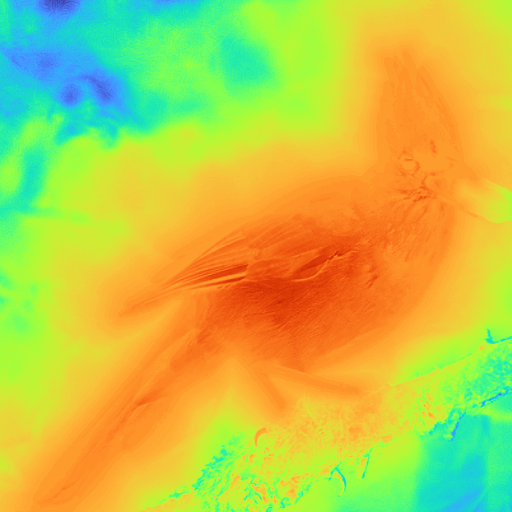}
    \caption{Heatmap}
  \end{subfigure}
\caption{Illustration of the SDF computation pipeline.}
\label{fig:imagesdfgeneration}
\end{figure}

\subsection{SDF Construction}

The process begins by extracting a robust mask of the salient object. Since traditional saliency detection often struggles to isolate semantic targets from cluttered backgrounds, we adopt a language-guided localization strategy. 
Specifically, we leverage Grounding DINO~\cite{liu2024grounding} for open-vocabulary detection. By integrating transformer-based architectures with grounded pre-training, Grounding DINO demonstrates superior proficiency in correlating linguistic descriptions with visual regions, allowing us to identify target objects via class labels with high semantic precision. The resulting bounding box acts as a geometric prompt for High Quality SAM (HQ-SAM)~\cite{ke2023segment}. Compared with the original Segment Anything Model (SAM), HQ-SAM introduces a learnable high-quality output token, which significantly enhances the object segmentation with complex structures or thin parts. This produces a fine-grained binary mask $I_{\mathrm{mask}}$ as in Fig.~\ref{fig:imagesdfgeneration} (b).
From this mask, we obtain the raw SDF by computing the signed difference between the inner and outer Euclidean Distance Transforms (DT):

\begin{equation}
  I_{\mathrm{SDF}} = \mathrm{DT}(I_{\mathrm{mask}}) - \mathrm{DT}(1 - I_{\mathrm{mask}})
  \label{eq:inner_outer}
\end{equation}

The inner transform $\mathrm{DT}(I_{\mathrm{mask}})$ (Fig.~\ref{fig:imagesdfgeneration} (c)) measures the minimum distance from object interior pixels to the boundary, while the outer transform $\mathrm{DT}(1 - I_{\mathrm{mask}})$ (Fig.~\ref{fig:imagesdfgeneration} (d)) encodes the distance from background pixels to the same boundary.
Within the finite lattice of the image domain, the inner and outer distance fields exhibit distinct spatial characteristics due to the inherent asymmetry between the object interior and the remaining background region. 
Consequently, these transforms yield distinct distance distributions with complementary spatial characteristics. This creates a pronounced numerical contrast between foreground and background regions and ensures clear structural differentiation across the image domain. 
To maintain stability during training, we normalize the raw SDF to match the scale of the pixel intensities in the input image, as shown in Fig.~\ref{fig:imagesdfgeneration} (f).

\subsection{Shape-guided Image Fusion}

The normalized SDF is integrated with the adversarial image to form a structurally enhanced representation $I_{\mathrm{fusion}}$ (Fig.~\ref{fig:imagesdfgeneration} (g)). The element-wise interaction enables the stable SDF signals to weight the importance of pixel-level appearance and accentuates object interiors while attenuating potentially deceptive adversarial noise. Through this integration, the network internalizes a balanced representation that leverages fine-grained textural cues for semantic precision while relying on geometric invariance for adversarial robustness.
The fusion is defined as:

\begin{equation}
I_{\mathrm{fusion}} = I_{\mathrm{adv}} \odot \left( 1 + \beta \, I_{\mathrm{SDF}} \right),
\label{eq8:img_enhance}
\end{equation}

\noindent where $\odot$ denotes element-wise multiplication and the  ratio $\beta \in [0, 1]$ controls the strength of geometric modulation. This mechanism adaptively amplifies object interior intensities while attenuating background regions, thereby reinforcing boundary contrast and guiding the model toward robust structural invariants. The effective incorporation of geometric cues into the fused representation is further validated via the difference heatmap in Fig.~\ref{fig:imagesdfgeneration} (h), which confirms that these cues are successfully embedded without distorting the semantic appearance.

\begin{algorithm}[tb]
\caption{GeoTexPuri Training Procedure}
\label{algo8:sdfgad}
\begin{algorithmic}[1]
\Statex \textbf{Notation:}
\Statex $f_\theta$: Trainable classifier with parameters $\theta$.
\Statex $\mathcal{A}(\cdot)$: Adversarial attack operator.
\Statex $\mathcal{S}(\cdot)$: SDF computation. 

\State \textbf{Input}: Dataset $\mathcal{D}$, Masks $\mathcal{M}$, Hyperparameter $\beta$.
\For{each step}
    \State Sample mini-batch: $(I_\mathrm{B}, y_\mathrm{B}) \sim \mathcal{D}$; Load masks: \(I_{\mathrm{mask}} \sim \mathcal{M}\)
    \State Generate adversarial samples: $I_{\mathrm{adv}} \leftarrow \mathcal{A}(I_\mathrm{B})$
   
    %\State // Geometric Guidance
    \State Generate SDF: $I_{\mathrm{SDF}} \leftarrow \mathcal{S}(I_{\mathrm{mask}})$
    \State Fusion: $I_{\mathrm{fusion}} = I_{\mathrm{adv}} \odot (1 + \beta \cdot I_{\mathrm{SDF}})$

    \State Loss: $L_\mathrm{total} = \mathcal{L}_{\mathrm{CE}}(f_\theta(I_\mathrm{B}), y) + \mathcal{L}_{\mathrm{CE}}(f_\theta(I_{\mathrm{adv}}), y) + \mathcal{L}_{\text{CE}}(f_{\theta}(I_{\mathrm{fusion}}), y)$
    
    \State Update $f_\theta$ via backpropagation
\EndFor
\end{algorithmic}
\end{algorithm}

\subsection{Training Objective}

The GeoTexPuri framework is optimized end-to-end through a multi-stream training strategy that facilitates the simultaneous learning of textural details and invariant geometric structures. By exposing the network to diverse input variations, we encourage the internalization of robust features that are less susceptible to pixel-level perturbations. Given a mini-batch of clean samples \((I_\mathrm{B}, y_\mathrm{B})\) and their corresponding adversarial counterparts \(I_{\mathrm{adv}}\), we define a composite objective function that incorporates standard classification losses alongside structural guidance.
The training process involves three parallel classification paths as Eq.~\ref{eq:totalsdfloss}. The first two paths focus on standard clean and adversarial classification using the cross-entropy loss $\mathcal{L}_{\text{CE}}$.
To explicitly reinforce geometric structure learning, we introduce a third processing branch that incorporates the shape-guided fusion image $I_{\mathrm{fusion}}$.

\begin{equation}
  L_{\mathrm{total}} = \mathcal{L}_{\text{CE}}(f_{\theta}(I_\mathrm{B}), y) + \mathcal{L}_{\text{CE}}(f_{\theta}(I_{\mathrm{adv}}), y) + \mathcal{L}_{\text{CE}}(f_{\theta}(I_{\mathrm{fusion}}), y)
  \label{eq:totalsdfloss}
\end{equation}

The total optimization objective $\mathcal{L}_{\text{total}}$ is the summation of these three loss components.
Through this multi-stream training, the model learns to maintain consistent predictive behavior even when the surface textures are distorted by adversarial noise. The overall procedure for GeoTexPuri is summarized in Algorithm~\ref{algo8:sdfgad}.

%------------------------------------------------------------------
%------------------------------------------------------------------

\begin{table}[tb]\setlength{\tabcolsep}{3pt} % 
\centering
\caption{The comparison of computation cost. The h denotes hours and m denotes minutes. RN denotes ResNet and CNX denotes ConvNeXt-L.}
\label{tab:computationcost}
\resizebox{1\textwidth}{!}{%
  \begin{tabular}{@{} l ccc ccc ccc @{}}
    \toprule
    & \multicolumn{3}{c}{Untargeted} & \multicolumn{3}{c}{Targeted} & \multicolumn{3}{c}{AutoAttack}\\
    \cmidrule(lr){2-4} \cmidrule(lr){5-7} \cmidrule(lr){8-10}
    Methods & Model & Train & Test & Model & Train & Test & Model & Train & Test  \\
    \midrule
    OSCP~\cite{lei2025instant} & RN50  &  71 h & 3 h & RN152 & 65h & 3h  & RN50 & 93h & 2h  \\
    GeoTexPuri & RN50 & 16.3h & 50 m  & RN152 & 22.8h & 40m  & RN50 & 19.5h & 35m\\ 
    GeoTexPuri & CNX-L & 61h & 10.8h & CNX-L  & 49h & 4.5h & CNX-L & 69h & 5h \\ 
    \bottomrule
  \end{tabular}
}

\end{table}

\section{Experiments}

\noindent\textbf{Datasets and Metrics.}
We split the ImageNet validation set into 40,000 images for training and 10,000 for testing, following OSCP~\cite{lei2025instant} to ensure a consistent comparison. All images are resized to a resolution of 512 \(\times\) 512.

\noindent\textbf{Implementation Details.}
We train and evaluate our framework on ResNet-50 under untargeted PGD-100 attacks and AutoAttack, ResNet-152 under targeted PGD-40 attacks. We also report the results of ConvNeXt-L model for all attacks. Each classifier is initialized with ImageNet-pretrained weights.
The initial learning rate is set to \(1\times10^{-4}\) and decayed using a StepLR scheduler with period 1250 and decay factor 0.5. The range of SDF pixel value is normalized to [-1, +1] and the shape control parameter \(\beta\) is fixed at 0.5 in all experiments.
We report clean accuracy, defined as performance on original unperturbed inputs, and robust accuracy, defined as performance on inputs with adversarial noise.

\noindent\textbf{Attack Setup.} 
Adversarial samples are generated using PGD~\cite{madry2017towards} and AutoAttack~\cite{croce2020reliable} under \(L_p\) constraints. 
We use \(\epsilon\) to denote the \(L_{\infty}\) perturbation bound, the maximum allowed per pixel change.
PGD-n denotes a PGD attack with n iterations.
For untargeted PGD-100, we set \(\epsilon=\) 4/255, step size \(\eta=\) 1/255, and 100 iterations.
For targeted PGD-40, we set \(\epsilon=\) 16/255, step size \(\eta=\) 0.4/255, and 40 iterations.
For AutoAttack, we use the standard ensemble consisting of four attacks applied in the following order: APGD with cross entropy loss, APGD with DLR loss, FAB, and Square, with \(\epsilon=\) 4/255.

\noindent\textbf{Computation Cost.}
The DINO+HQ-SAM masks are precomputed once as an offline cost of approximately 48 hours and do not contribute to the training time.
Masks are stored as binary files rather than SDF float values for I/O efficiency. The SDF is calculated during training at negligible additional cost. 
As in Table~\ref{tab:computationcost}, GeoTexPuri trains for 16.3 hours on a single A100 GPU with 50 minutes inference under untargeted PGD-100 attack.
All experiments under AutoAttack are conducted on a faster H100 GPU for our framework and on 4 GPUs for OSCP. GeoTexPuri reduces training time from 93 hours to 19.5 hours and inference time from 2 hours to 35 minutes. The reduced computational cost also makes training with larger backbones ConvNeXt-L practically feasible.

\begin{table}[tb]\setlength{\tabcolsep}{6pt} 
\centering
 \caption{Comparison with state-of-the-art adversarial defense methods on ImageNet under three attack settings. Clean and robust accuracies are reported along with the classifier used by each method.}
  \label{tab:comparesdfsota}
  \begin{tabular}{@{}llccc@{}}
    \toprule
    Attacks & Defense & Clean Acc & Robust Acc & Classifiers \\
    \midrule
    \multirow{5}{*}{Untargeted} & Without defense & 80.55 & 0.01 & ResNet-50 \\
    & GDMP~\cite{wang2022guided} & 73.53 & 72.97 & ResNet-50  \\
    & OSCP~\cite{lei2025instant} & 77.63 & 73.89 & ResNet-50 \\
    \cmidrule{2-5}
    & GeoTexPuri (Ours) & 77.57 & 74.47 & ResNet-50 \\
    & GeoTexPuri (Ours) & \textbf{82.87} & \textbf{81.51} & ConvNeXt-L \\
    \midrule
    \multirow{4}{*}{Targeted} & Without defense & 82.33 & 0.04 & ResNet-152  \\
    & GDMP~\cite{wang2022guided} & 78.10 & 77.86 & ResNet-152 \\
    & OSCP~\cite{lei2025instant} & 79.81 & 78.78 & ResNet-152 \\
    \cmidrule{2-5}
    & GeoTexPuri (Ours) & 81.55 & 80.90 & ResNet-152 \\
    & GeoTexPuri (Ours) & \textbf{84.31} & \textbf{83.38} & ConvNeXt-L \\
    \midrule
    \multirow{7}{*}{AutoAttack} & Without defense & 80.55 & 0.00 & ResNet-50 \\
    & MeanSparse~\cite{amini2024meansparse} & 77.96 & 59.64 & ConvNeXt-L \\
    & Singh~\cite{singh2023revisiting} & 77.00 & 57.70 & ConvNeXt-L \\
    & DiffPure~\cite{nie2022diffusion} & 75.77 & 73.02 & ResNet-50 \\
    & OSCP~\cite{lei2025instant} & 77.63 & 74.19 & ResNet-50 \\ 
    \cmidrule{2-5}
    & GeoTexPuri (Ours) & 78.25 & 75.47 & ResNet-50 \\
    & GeoTexPuri (Ours) & \textbf{84.79} & \textbf{83.52} & ConvNeXt-L \\
  \bottomrule
  \end{tabular}
\end{table}

\subsection{Comparison with State-of-the-art}
To assess the performance of GeoTexPuri, we present a comprehensive comparison with state-of-the-art defense mechanisms on ImageNet, as summarized in Table~\ref{tab:comparesdfsota}. 
The results show that our approach consistently outperforms existing methods, achieving superior clean and robust accuracy across various adversarial settings.
Under the untargeted PGD-100 attack, our framework achieves 74.47\% robust accuracy using the ResNet-50 backbone, which exceeds the leading diffusion-based defense OSCP~\cite{lei2025instant}. In the targeted PGD-40 setting, GeoTexPuri improves performance to 81.55\% clean and 80.90\% robust accuracy using the same ResNet-152 backbone.
For the rigorous AutoAttack benchmark, our framework improves clean accuracy by 6.83\% (84.79\% vs. 77.96\%) and robust accuracy by 9.33\% (83.52\% vs. 74.19\%) utilizing a ConvNeXt-L backbone. 
%Notably, GeoTexPuri is the first defense framework to exceed the 80\% robust accuracy threshold on ImageNet under AutoAttack, which highlights the critical importance of balancing texture with invariant geometric structures. 
%These results confirm that our deterministic approach provides a more effective and efficient defense than stochastic generative purification methods.
In addition, we aim to establish a more challenging benchmark by adopting larger models and stronger attacks. ConvNeXt-L has been adopted in recent adversarial training works, but remains unexplored in adversarial purification due to its prohibitive training cost. Our framework makes this feasible.

\begin{figure}[tb]
  \centering
  \begin{subfigure}{0.98\linewidth}
    \includegraphics[width=\linewidth]{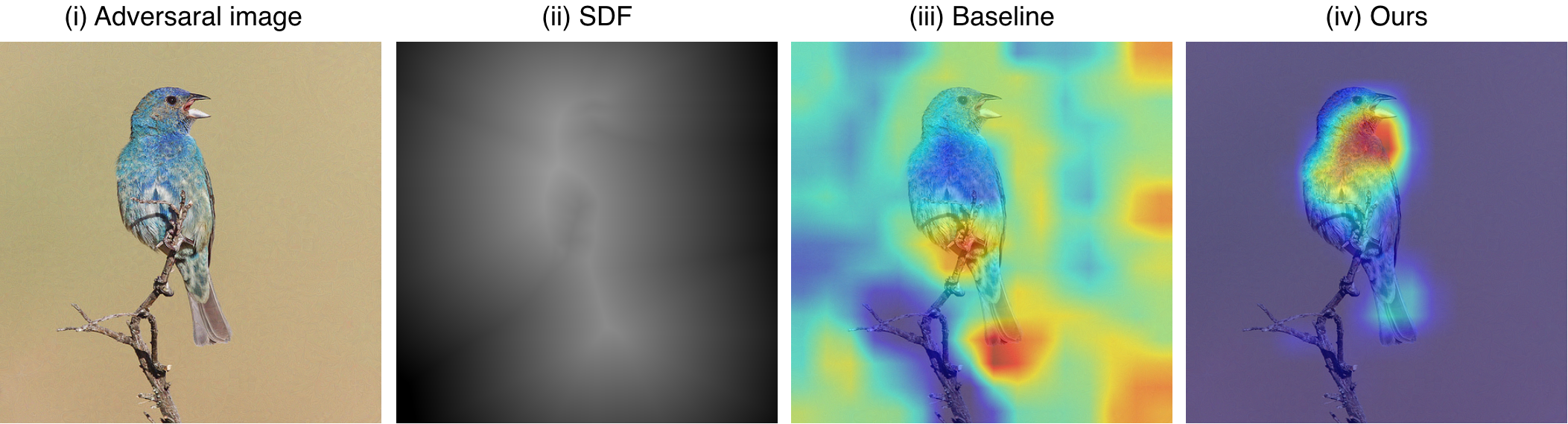}
    \caption{Under untargeted PGD-100 attack.}
  \end{subfigure}
  \begin{subfigure}{0.98\linewidth}
    \includegraphics[width=\linewidth]{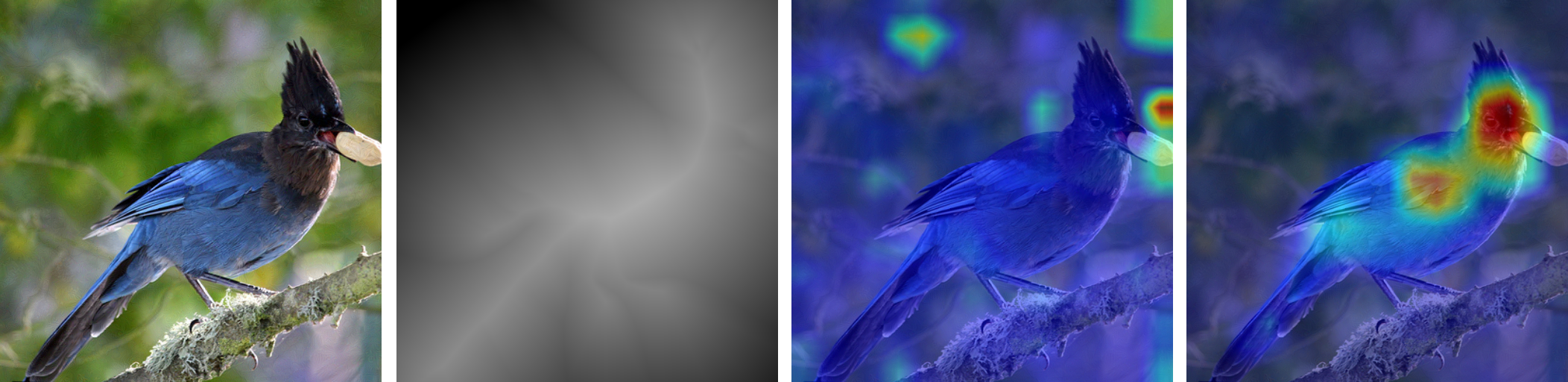}
    \caption{Under targeted PGD-40 attack.}
  \end{subfigure}
  \begin{subfigure}{0.98\linewidth}
    \includegraphics[width=\linewidth]{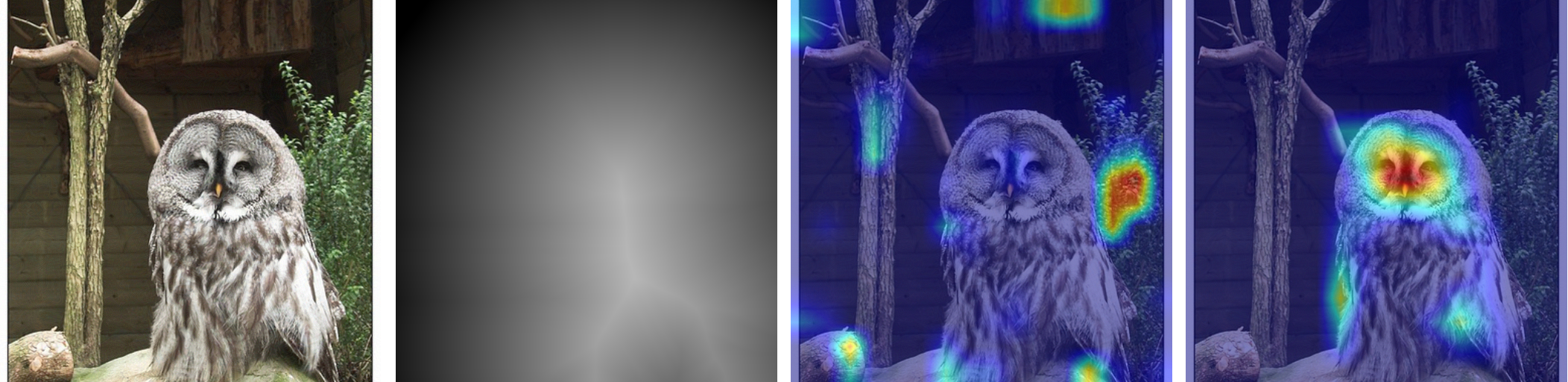}
    \caption{Under AutoAttack.}
  \end{subfigure}
  \caption{Comparison of Grad-CAM pseudo-color heatmaps. To provide clear spatial localization, these maps are overlaid on the images.  From left to right, the columns present: (i) Adversarial image, (ii) its corresponding SDF field, (iii) the Grad-CAM without defense, and (iv) the Grad-CAM of our GeoTexPuri framework. Heatmap colors represent feature importance, with red denoting peak activation and blue denoting background or noise.}
  \label{fig:gradcam}
\end{figure}

\noindent\textbf{Qualitative Analysis via Grad-CAM Interpretability.}
To interpret the underlying mechanism of GeoTexPuri robustness, we use Grad-CAM~\cite{selvaraju2017grad}, which generates gradient-weighted class activation maps to visualize decision-making regions through a pseudo-color heatmap as illustrated in Fig.~\ref{fig:gradcam}.
The red signifies peak activation of core semantic features, while blue and violet denote ignored background components or noise. The baseline models without defense exhibit fragmented and chaotic activation patterns under all adversarial scenarios. The high-importance (red) regions frequently drift away from the target object and dissipate into the background.
In contrast, GeoTexPuri demonstrates a strong ability to maintain a concentrated and consistent focus on the object through invariant structure. 
%This stable activation pattern proves that our multi-stream training framework effectively encourages the network to disregard deceptive adversarial noise even when the SDF is not provided during inference. 

\subsection{Ablation Study}

\begin{table}[tb]\setlength{\tabcolsep}{5pt} % 
\centering
\caption{Comprehensive ablation studies on ImageNet. Bold indicates the best performance within each group. For untargeted PGD-100, both train and test attack classifiers are ResNet-101. For targeted PGD-40, ResNet-152 is used. For AutoAttack, ConvNeXt-L is used.}
\label{tab:allablation}
\resizebox{1\textwidth}{!}{%
  \begin{tabular}{@{} l l cc cc cc @{}}
    \toprule
    & & \multicolumn{2}{c}{Untargeted PGD-100} & \multicolumn{2}{c}{Targeted PGD-40} & \multicolumn{2}{c}{AutoAttack} \\
    \cmidrule(lr){3-4} \cmidrule(lr){5-6} \cmidrule(lr){7-8}
    Group & Setting & Clean & Robust & Clean & Robust & Clean & Robust \\
    \midrule
    
    \multirow{4}{*}{Shapes} 
    & Edge     & 77.79 & 75.96 & 80.48 & 80.07 & 83.77 & 82.01 \\
    & Contour  & 79.34 & 76.67 & 80.73 & 80.11 & 84.50 & 82.84 \\
    & Skeleton & 79.38 & 76.30 & 80.85 & 80.46 & 84.53 & 83.13 \\
    & SDF (Ours) & \textbf{80.00} & \textbf{77.35} & \textbf{81.55} & \textbf{80.90} & \textbf{84.79} & \textbf{83.52} \\
    
    \midrule
    \multirow{3}{*}{Masks}
    & Otsu     & 79.47 & 76.43 & 80.73 & 79.02 & 84.06 & 81.64 \\
    & InSPyReNet~\cite{kim2022revisiting} & 79.73 & 77.13 & 81.27 & 80.40 & \textbf{85.10} & 83.41 \\
    & DINO (Ours) & \textbf{80.00} & \textbf{77.35} & \textbf{81.55} & \textbf{80.90} & 84.79 & \textbf{83.52} \\
    \midrule
    \multirow{4}{*}{\shortstack[l]{Bounds\\$\epsilon_{\infty}$}}
    & 2/255  & \textbf{80.15} & \textbf{77.67} & \textbf{82.18} & \textbf{81.65} & \textbf{84.97} & \textbf{83.67} \\
    & 4/255  & 80.00 & 77.35 & 81.95 & 81.19 & 84.79 & 83.52 \\
    & 8/255  & 78.17 & 74.95 & 81.62 & 80.57 & 84.57 & 83.17 \\
    & 16/255 & 76.34 & 73.14 & 81.55 & 80.90 & 84.46 & 82.87 \\
    
    \midrule
    \multirow{5}{*}{\shortstack[l]{Ratios\\$\beta$}}
    & 0.3 & 79.44 & 77.23 & 81.13 & 80.95 & 83.96 & 82.89 \\
    & 0.4 & 79.71 & 76.89 & 81.63 & 81.15 & 84.22 & 82.96 \\
    & 0.5 & \textbf{80.00} & \textbf{77.35} & 81.55 & 80.90 & \textbf{84.79} & \textbf{83.52} \\
    & 0.6 & 79.86 & 76.97 & 81.74 & 81.25 & 84.00 & 82.83 \\
    & 0.7 & 79.52 & 76.76 & \textbf{81.82} & \textbf{81.69} & 84.15 & 83.01 \\
    \bottomrule
  \end{tabular}
}
\end{table}

\begin{figure}[tb]
  \centering
  \begin{subfigure}{0.24\linewidth}
    \includegraphics[width=\linewidth]{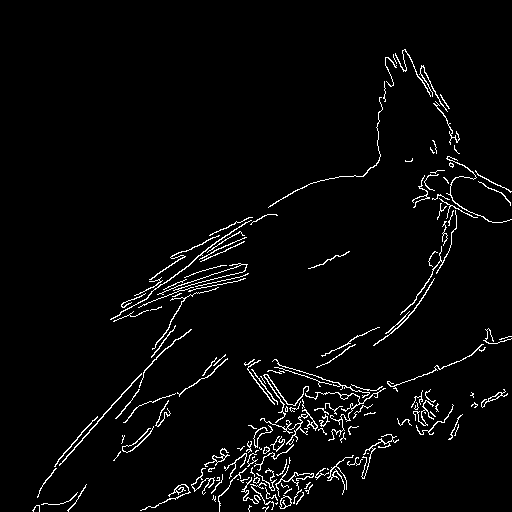}
    \caption{Edge}
  \end{subfigure}
  \begin{subfigure}{0.24\linewidth}
    \includegraphics[width=\linewidth]{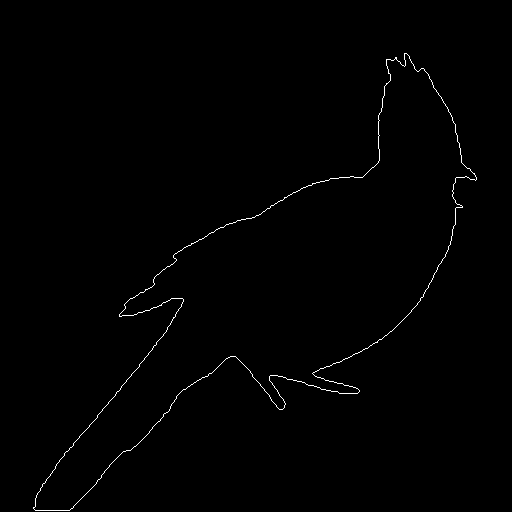}
    \caption{Contour}
  \end{subfigure}
  \begin{subfigure}{0.24\linewidth}
    \includegraphics[width=\linewidth]{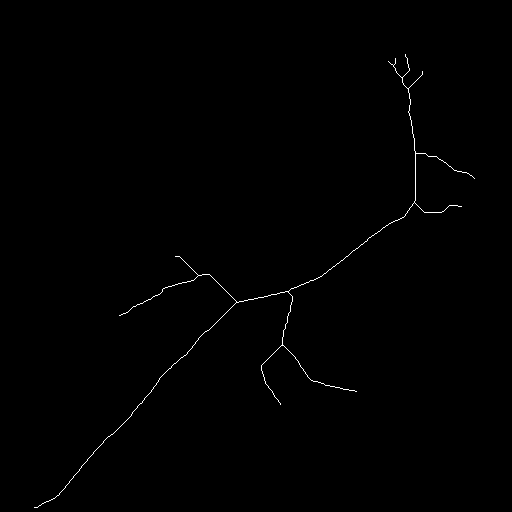}
    \caption{Skeleton}
  \end{subfigure}
  \begin{subfigure}{0.24\linewidth}
    \includegraphics[width=\linewidth]{Figures/sdf/175_sdf.png}
    \caption{SDF}
  \end{subfigure}
  \caption{Visualization of diverse geometric representations.}
  \label{fig:variousshapes}
\end{figure}

\noindent\textbf{Comparative Analysis of Geometric Representation.} 
We evaluate the efficacy of the spacial field SDF by comparing it with three alternative geometric representations: edges, contours, and skeletons, as in the \textit{Shapes} group of Table~\ref{tab:allablation}. For edge encoding, the Canny operator is employed to extract object boundaries. 
Contour extraction and skeletonization utilize the same high-quality mask as the SDF. 
%Specifically, contour extraction delineates the external silhouette by tracing continuous boundary curves along foreground-background transitions, while skeletonization reduces the mask to its topological medial axis to capture the underlying structural connectivity as a one-pixel-wide representation.
As illustrated in Fig.~\ref{fig:variousshapes}, each method captures distinct structural aspects of the object.
The results in Table~\ref{tab:allablation} demonstrate that GeoTexPuri achieves consistent robustness gains across all geometric types, which indicates that the framework can leverage diverse structural cues to counteract adversarial noise without strict dependence on any particular geometric prior. Among these candidates, the SDF outperforms the other three geometric encodings and is therefore adopted as the default configuration.

\noindent\textbf{Comparative Analysis of Mask Generator.} 
To analyze how different mask generation strategies influence performance, we compare our proposed pipeline with two alternative designs, as reported in the \textit{Masks} group of Table~\ref{tab:allablation}.
The first approach generates a binary mask using Otsu thresholding, followed by connected component analysis to select the principal object. 
Although this method produces relatively coarse binary masks, the resulting framework achieves 79.47\% clean and 76.43\% robust accuracy under untargeted PGD-100 attack. The gap between Otsu and DINO+HQ-SAM is only 0.92\%, which suggests that GeoTexPuri framework does not strictly rely on high-quality masks and remains robust even with sub-optimal mask quality.
In the second configuration, we employ a pre-trained salient object detection model, InSPyReNet~\cite{kim2022revisiting}, to generate the object masks. Benefiting from more accurate object localization, this strategy leads to slightly performance improvement over simple thresholding.
Finally, our approach leverages Grounding DINO~\cite{liu2024grounding} to localize target objects and applies HQ-SAM~\cite{ke2023segment} to produce high-quality segmentation masks. This combination achieves the highest robust accuracy.

\noindent\textbf{Sensitivity Analysis of Perturbation Bound \(\epsilon_{\infty}\).}
We investigate the stability of our framework across a range of perturbation bounds $\epsilon_{\infty}$, as shown in the \textit{Bounds} group of Table~\ref{tab:allablation}. While a slight performance decay is observed as $\epsilon_{\infty}$ increases from 2/255 to 16/255, the framework maintains a high level of accuracy throughout. The defense remains remarkably stable even under a significant attack magnitude of $\epsilon_{\infty} = $ 16/255. 
This stable performance indicates that GeoTexPuri effectively decouples semantic structures from adversarial noise, enabling robust inference across a wide range of threat levels.

\noindent\textbf{Sensitivity Analysis of Fusion Ratio $\beta$.}
We evaluate various ratios from 0.3 to 0.7 and show the results in the bottom \textit{Ratios} group of Table~\ref{tab:allablation}. The model achieves its peak performance when $\beta$ is set to 0.5 under the untargeted and AutoAttack. 
For targeted attacks, the best performance reaches at $\beta=$ 0.7 with 81.82\% clean and 81.69\% robust accuracy. 
Despite the optimal value shifts, the framework demonstrates low sensitivity to the fusion ratio as performance fluctuations falling within a narrow 1\%-2\% margin.
Considering the trade-offs across attacks, we adopt $\beta=$ 0.5 as the default value for our final model.

\noindent\textbf{Comparative Analysis of Classifiers.}
To evaluate the robustness of our framework, we select representative backbones from distinct architectural categories. Specifically, we  include the classic CNN-based ResNet variants~\cite{he2016deep} (50, 101, and 152), the prominent vision transformer Swin-B~\cite{liu2021swin}, and the large-scale ConvNeXt-L~\cite{liu2022convnet}.
To disentangle the effect of model capacity from framework contribution, we conduct a comprehensive analysis across all combinations of training and test attack classifiers under untargeted PGD-100 in Table~\ref{tab:ablationmodel}. 
%The column denotes the classifier used to generate adversarial examples during training and the row denotes the attack classifier used during test. When a smaller classifier ResNet-50 is used during training, employing larger classifiers at test time does not lead to accuracy degradation. Conversely, models trained with larger classifiers ConvNeXt-L retain their robustness when evaluated against smaller ones. 
These results confirm that the robustness gains are attributable to the framework and remain stable across diverse training and test classifier configurations.

\begin{table}[tb]%\setlength{\tabcolsep}{5pt} % 
\centering
\caption{Cross-backbone evaluation under untargeted PGD-100 attack.}
\label{tab:ablationmodel}
\resizebox{1\textwidth}{!}{%
  \begin{tabular}{@{} ll cc cc cc cc cc @{}}
    \toprule
    & \multirow{2}{*}{\diagbox{Train}{Test}}  & \multicolumn{2}{c}{ResNet-50} & \multicolumn{2}{c}{ResNet-101} & \multicolumn{2}{c}{ResNet-152} & \multicolumn{2}{c}{Swin-B} & \multicolumn{2}{c}{ConvNeXt-L}\\
    \cmidrule(lr){3-4} \cmidrule(lr){5-6} \cmidrule(lr){7-8} \cmidrule(lr){9-10} \cmidrule(lr){11-12}
    Methods && Clean & Robust & Clean & Robust & Clean & Robust & Clean & Robust & Clean & Robust \\
    \midrule
    OSCP~\cite{lei2025instant} & ResNet-50  & 77.63 & 73.89 & 72.57 & 69.05 & 72.41 & 69.04 & 71.99 & 69.76 & 73.15 & 69.54 \\
    \midrule
    \multirow{5}{*}{GeoTexPuri} &  ResNet-50  & 77.57 & 74.47 & 77.51 & 75.92 & 77.63 & 76.03 & 77.59 & 76.67 & 77.45 & 75.80  \\
    & ResNet-101 & 79.83 & 79.04 & 80.00 & 77.35 & 79.91 & 78.75 & 79.82 & 79.20 & 79.70 & 78.51 \\
    & ResNet-152 & 81.74 & 79.92 & 80.83 & 79.55 & 80.70 & 78.07 & 80.75 & 80.18 & 80.77 & 79.60 \\
    & Swin-B & 82.32 & 81.70 & 82.27 & 81.38 & 82.20 & 81.06 & 82.36 & 80.61 & 82.29 & 81.01 \\  
    & ConvNeXt-L & 82.93 & 82.42 & 82.98 & 82.12 & 82.84 & 82.11 & 82.69 & 82.37 & 82.87 & 81.51  \\
    \bottomrule
  \end{tabular}
}
  
\end{table}

\noindent\textbf{Performance of Full Imagenet Dataset.}
To further demonstrate the stability and scalability of our approach, we extend GeoTexPuri to the full ImageNet-1k dataset, which consists of 1,281,167 training images and 50,000 validation images across 1,000 categories, as in Table~\ref{tab:fullimagenet}. 
%To ensure a rigorous comparison, we maintain the same attack configurations as in our previous experiments. 
Our framework achieves a clean accuracy of 81.39\% and a robust accuracy of 80.54\% under targeted PGD-40. 
Due to the large parameter scale and computational cost of ConvNeXt-L, we switch the backbone to ResNet-101 for AutoAttack, achieving a robust accuracy of 79.83\%.
These results confirm that GeoTexPuri remains stable and effective when scaled to million-level training samples.

\begin{table}[tb]\setlength{\tabcolsep}{6pt}
\centering
\caption{Performance on full ImageNet dataset. (Accuracy in \%)}
\label{tab:fullimagenet}
  \begin{tabular}{@{}l cccccc@{}}
    \toprule
    & \multicolumn{2}{c}{Untargeted PGD-100} & \multicolumn{2}{c}{Targeted PGD-40} & \multicolumn{2}{c}{AutoAttack} \\
    \cmidrule(lr){2-3} \cmidrule(lr){4-5} \cmidrule(lr){6-7}
    Model & Clean & Robust & Clean & Robust & Clean & Robust \\
    \midrule
    ResNet-101 & 79.80 & 79.72 & ---   & ---   & 78.65 & 79.83 \\
    ResNet-152 & ---   & ---   & 81.39 & 80.54 & ---   & ---   \\
    \bottomrule
  \end{tabular}

\end{table}

\begin{figure}[tb]
  \centering
  \begin{subfigure}{0.24\linewidth}
    \includegraphics[width=\linewidth]{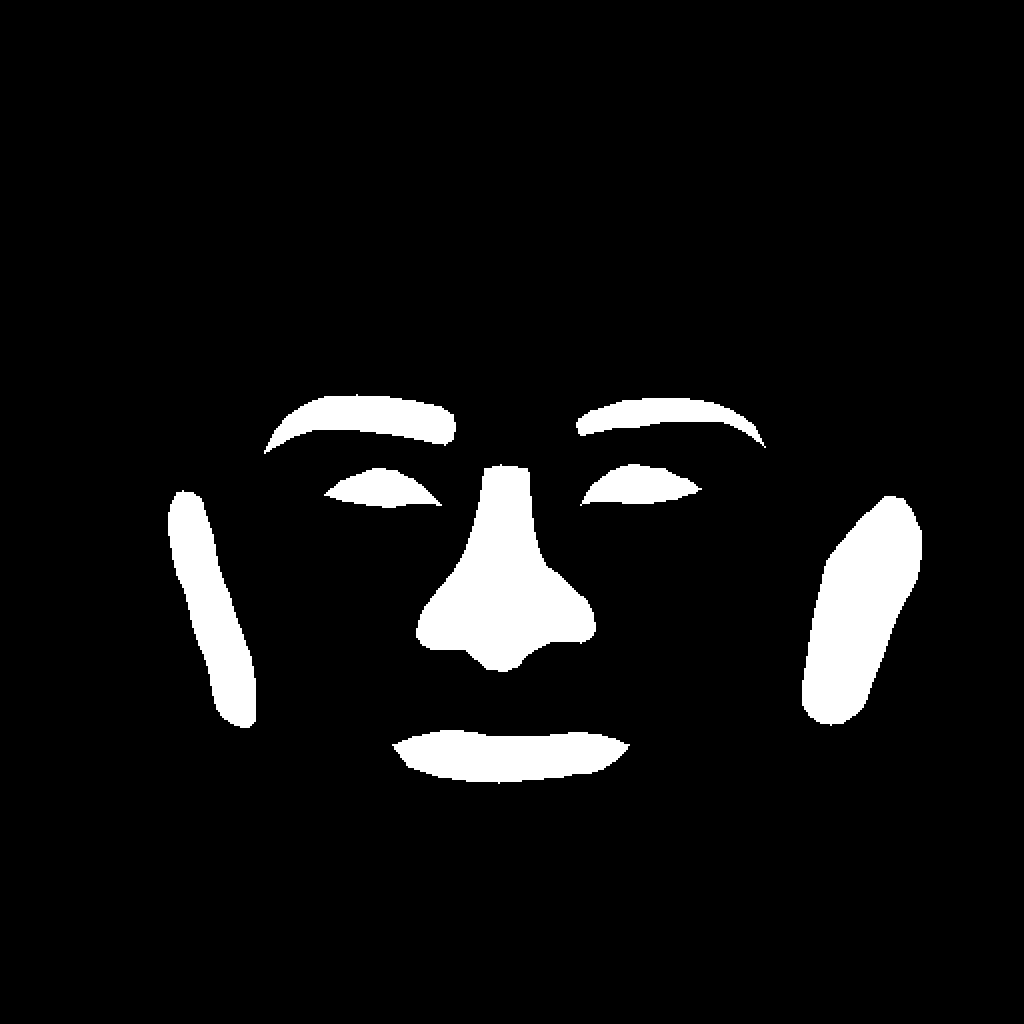}
    \caption{Object Mask}
  \end{subfigure}
  \begin{subfigure}{0.24\linewidth}
    \includegraphics[width=\linewidth]{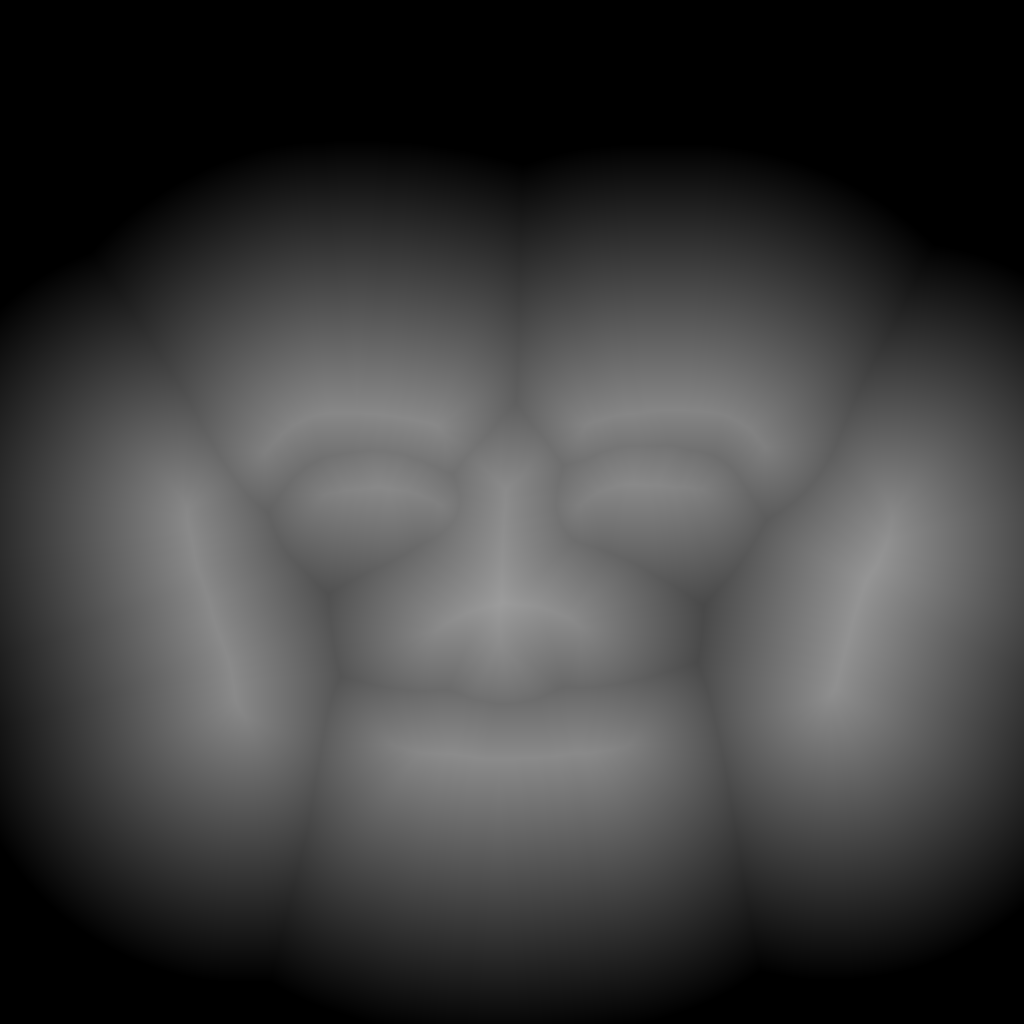}
    \caption{SDF}
  \end{subfigure}
  \begin{subfigure}{0.24\linewidth}
    \includegraphics[width=\linewidth]{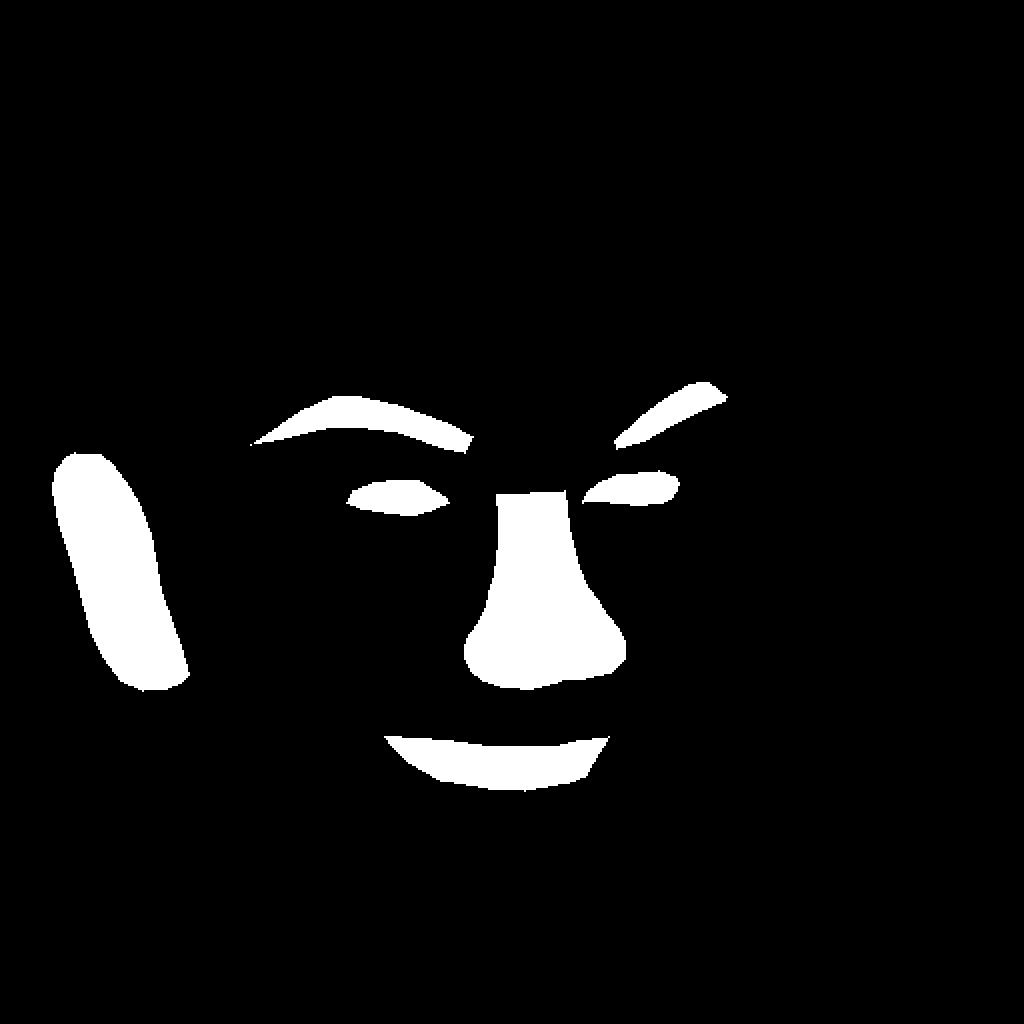}
    \caption{Object Mask}
  \end{subfigure}
  \begin{subfigure}{0.24\linewidth}
    \includegraphics[width=\linewidth]{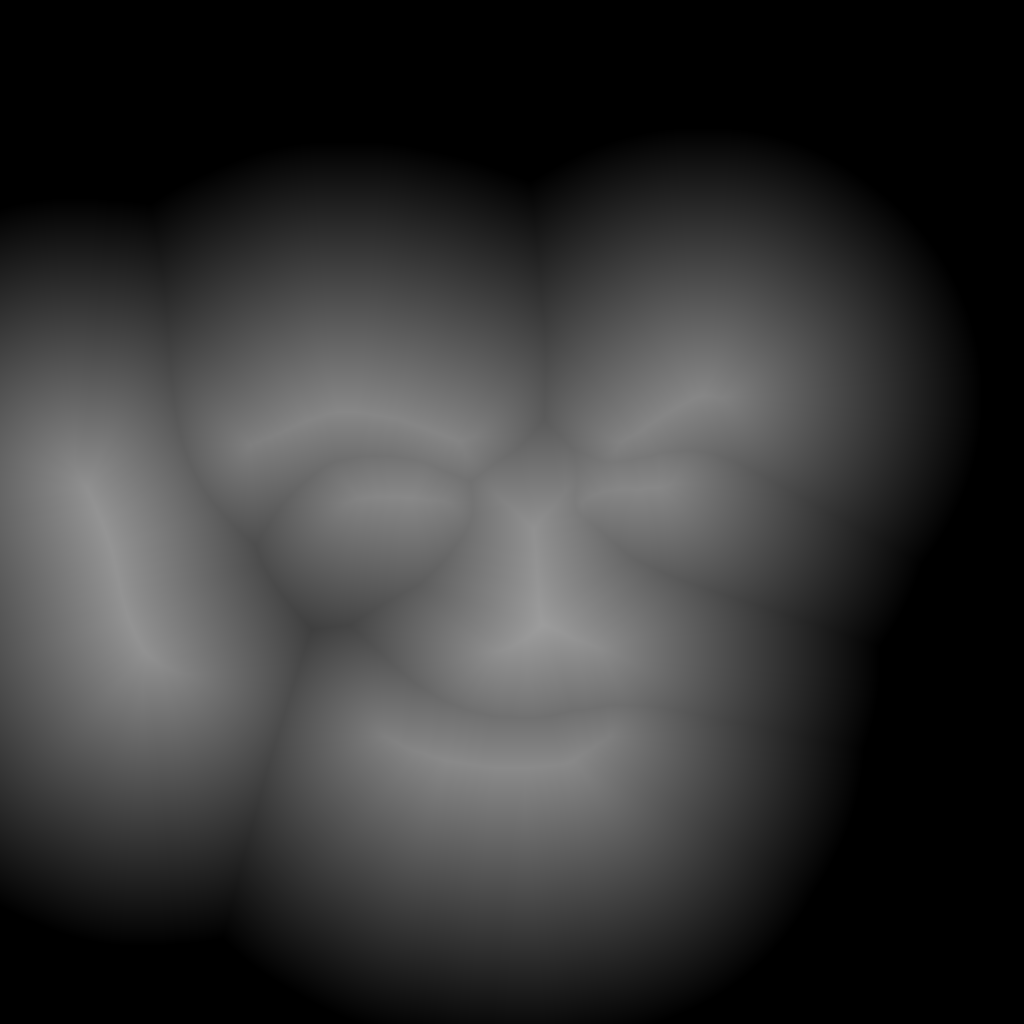}
    \caption{SDF}
  \end{subfigure}
  \caption{Visualizations of unified facial masks and their corresponding SDF across different samples.}
  \label{fig:facemasksdf}
\end{figure}

\noindent\textbf{Performance of CelebA-HQ Dataset.}
To evaluate the efficacy of our framework to face recognition task, we extend our study to the CelebA-HQ~\cite{liu2015deep} dataset using an ArcFace~\cite{deng2019arcface} backbone. This setup involves 28,299 images (23,533 training and 4,766 test samples), with semantic masks derived from CelebAMask-HQ~\cite{lee2020maskgan}. 
We construct a unified mask (Fig.~\ref{fig:facemasksdf} (a) (c)) by combining key facial components, such as eyes, brows, ears, nose, and mouth, to calculate the SDF (Fig.~\ref{fig:facemasksdf} (b) (d)) for training.
As shown in Table~\ref{tab:faceacc}, we evaluate the model under targeted attacks across three parameter groups $(\epsilon_{\infty}, \text{iterations}, \eta)$. 
%Under standard constraints (4/255, 10, 0.25/255), the system achieves a clean and robust accuracy of 88.57\% and 87.21\%. Interestingly, when increasing the attack intensity to (4/255, 100, 1/255), the robust accuracy improves slightly to 87.89\% despite a minor drop in clean accuracy. Most notably, under a larger perturbation budget of 16/255, the model exhibits exceptional stability with a clean and robust accuracy of 88.48\% and 95.80\%. 
These results demonstrate that our approach effectively safeguards identity features and establishes high defensive robustness under varying adversarial intensities in face recognition tasks.

\begin{table}[tb]\setlength{\tabcolsep}{15pt} 
\centering
\caption{Performance on CelebA-HQ dataset with different attack parameters.} 
\label{tab:faceacc}
  \begin{tabular}{@{} ccc cc @{}}
    \toprule
    \multicolumn{3}{c}{Hyper-parameters} & \multicolumn{2}{c}{Accuracy (\%)} \\
    \cmidrule(r){1-3} \cmidrule(l){4-5}
    $\epsilon_{\infty}$ & Iterations & Step size $\eta$ & Clean & Robust \\
    \midrule
    4/255  & 10  & 0.25/255 & 89.51 & 92.74 \\
    4/255  & 100 & 1/255    & 86.43 & 87.89 \\
    16/255 & 40  & 0.4/255  & 88.48 & 95.80 \\
    \bottomrule
  \end{tabular}
\end{table}

\section{Conclusion}
In this paper, we propose Geometric and Texture balancing Purification (GeoTexPuri), a novel defensive framework that harmonizes invariant geometric structures with textural features to enhance adversarial robustness. By integrating Signed Distance Field (SDF) as geometric guidance into a multi-stream training phase, our approach enables neural networks to internalize purified representations that effectively decouple semantic structures from adversarial perturbations.
Comprehensive evaluations on ImageNet demonstrate the superior efficacy of our method compared to existing state-of-the-art defenses.
Notably, GeoTexPuri achieves a robust accuracy of 83.52\% under the rigorous AutoAttack benchmark while maintaining high clean accuracy. Furthermore, our framework remains highly efficient at inference time because it operates as a standard classifier without requiring auxiliary geometric inputs. These results suggest that incorporating geometric priors is a promising direction for developing robust, scalable, and efficient defense systems.

%\clearpage\mbox{}Page \thepage\ of the manuscript. 
\par\vfill\par

\clearpage  % TODO FINAL: This \clearpage needs to be removed from both review and camera-ready versions.

\section*{Acknowledgements}
High-performance computing resources were provided by the Erlangen National High Performance Computing Center (NHR@FAU) at FAU Erlangen-Nürnberg (FAU), under the NHR projects b143dc and b180dc. NHR is funded by federal and Bavarian state authorities, and NHR@FAU hardware is partially funded by the German Research Foundation (DFG) – 440719683. Additional support was  received by the ERC - project MIA-NORMAL 101083647,  DFG 513220538, 512819079, and by the state of Bavaria (HTA).

\bibliographystyle{splncs04}
\bibliography{main}
\end{document}